\documentclass{article}
\usepackage{spconf,amsmath,graphicx,diagbox,makecell,marvosym}

\title{Neural network fragile watermarking with no model performance degradation}
\name{Zhaoxia Yin$^{1}$, Heng Yin$^{2}$, Xinpeng Zhang$^{3}$\thanks{Copyright 2022 IEEE. Published in 2022 IEEE International Conference on Image Processing (ICIP), scheduled for 16-19 October 2022 in Bordeaux, France. Personal use of this material is permitted. However, permission to reprint/republish this material for advertising or promotional purposes or for creating new collective works for resale or redistribution to servers or lists, or to reuse any copyrighted component of this work in other works, must be obtained from the IEEE. Contact: Manager, Copyrights and Permissions / IEEE Service Center / 445 Hoes Lane / P.O. Box 1331 / Piscataway, NJ 08855-1331, USA. Telephone: + Intl. 908-562-3966. This work is supported by National Natural Science Foundation of China under Grant No.62172001, U1936214.}}
\address{$^{1}$School of Communication and Electronic Engineering,\\East China Normal University,
Shanghai, China\\
$^{2}$ Anhui Provincial Key Laboratory of Multimodal Cognitive Computation,  Anhui University\\
$^{3}$ School of Computer Science and Technology, Fudan University, Shanghai, China\\
$^{1}$zxyin@cee.ecnu.edu.cn;$^{2}$e21301335@stu.ahu.edu.cn;$^{3}$zhangxinpeng@fudan.edu.cn}
\begin{document}
%

\maketitle
	\begin{abstract}
		
		Deep neural networks are vulnerable to malicious fine-tuning attacks such as data poisoning and backdoor attacks. 
		Therefore, in recent research, it is proposed how to detect malicious fine-tuning of neural network models. 
		However, it usually negatively affects the performance of the protected model. 
		Thus, we propose a novel neural network fragile watermarking with no model performance degradation. 
		In the process of watermarking, we train a generative model with the specific loss function and secret key to generate triggers that are sensitive to the fine-tuning of the target classifier. 
		In the process of verifying, we adopt the watermarked classifier to get labels of each fragile trigger. 
		Then, malicious fine-tuning can be detected by comparing secret keys and labels. 
		Experiments on classic datasets and classifiers show that the proposed method can effectively detect model malicious fine-tuning with no model performance degradation.
	\end{abstract}
	\section{Introduction}
	\label{sec:intro}
	The performance of Deep Neural Network (DNN) in image
    recognition \cite{krizhevsky2012imagenet,he2016deep,dosovitskiy2020image}, natural language processing \cite{collobert2008unified,luong2015effective,kenton2019bert}, speech recognition \cite{dahl2011context} and explainable machine learning \cite{samek2019towards,mathews2019explainable} has achieved excellent results. 
    Meanwhile, the requirement of powerful computing resources and a long training time in obtaining a business model renders many users directly download the pre-trained DNN models from websites or use enterprise cloud-based services. 
    However, these pre-trained models may have been injected backdoor by maliciously fine-tuning them to be backdoor models. 
	It may cause a severe security accident when users unknowingly apply the backdoor model to applications like autonomous driving. 
	Hence, we resort to the technology of model watermarking to verify the integrity of pre-trained models.
    \par
	The mainstream model watermarking technologies are currently divided into two types. 
	One is to embed the watermarking by modifying parameters of the model, like \cite{guan2020reversible,botta2021neunac}, and the other is to implement the trigger set, like \cite{he2019sensitive,xu2020secure,zhu2021fragile}.
	Robustness is one feature of model watermarking that we cared about most, which means that model watermarking should be robust to model modification attacks such as fine-tuning and compression. 
	Fragility is opposite to robustness, which means that watermarking is sensitive to model modifications. 
	Modifications of the watermarked model can be detected immediately as long as it undergoes fine-tuning. 
    \par
	Let's review recent neural network fragile watermarking approaches for detecting malicious fine-tuning.
	In \cite{guan2020reversible}, a reversible fragile watermarking scheme for model integrity authentication is proposed. 
	This scheme utilizes the pruning theory of model compression technology to construct a host sequence. 
	Then, it adopts histogram shift \cite{ni2006reversible} technology to embed watermarking information.
    In \cite{botta2021neunac}, a non-reversible algorithm is applied to embed the watermarking in a secret frequency domain defined by a linear transformation, this scheme has no degradation in performances of the watermarked neural network with respect to the original one. 
	However, research works \cite{guan2020reversible,botta2021neunac} are white-box fragile watermarking methods that need the detail of networks, which is difficult to verify integrity of watermarked models remotely.  
	In \cite{he2019sensitive}, fragile triggers are generated by adding a small perturbation to the original training data, and this method first uses designed transformed inputs as a defense to protect the integrity property of DNN. 
	Compared with \cite{he2019sensitive}, research work \cite{xu2020secure} transforms all complex activation functions into polynomials to facilitate the generation of fragile triggers which apply to all neural network frameworks. 
	Though research works \cite{he2019sensitive,xu2020secure} are black-box fragile watermarking methods that are convenient to verify model integrity remotely, the generation of fragile triggers still needs the parameters of target networks.
	\cite{zhu2021fragile}-KSEM 2021 puts forward an alternate two-stage training strategy with specific loss to embed the generated trigger set into target model. 
	Though this black-box method can detect malicious fine-tuning sensitively, it needs to fine-tune the target classifier to embed the trigger set, and the performance of the watermarked classifier also gets degradation. 
	Therefore, we propose a novel fragile watermarking method to mark the target classifier by generating the fragile trigger set from a generative model trained with specific loss and a secret key. 
	Watermarking process of the target classifier and training process of the generative model happen simultaneously. 
	No detail and modification of the target classifier is required, and we only need to query the target classifier in the whole watermarking process.
	\par
	Our contributions in model fragile watermarking are summarized as follows:
		\begin{itemize}
		\item A novel black-box method is proposed to design the fragile trigger set for verifying the integrity of watermarked classifier.  
		\item A regularization term $Var$ is first utilized in training generators, we illustrate that $Var$ is the key for generating a fragile trigger set in Sect 4.1.
		\item The fragile trigger set can perceive the modification of watermarked classifier in the first epoch of fine-tuning stage, which is demonstrated in Sect 4.2.
		\item Our fragile watermarking method is compatible and effective for classifiers with increasing parameters and datasets with more categories. Corresponding experiment results are given in Sect 4.3.
	\end{itemize} 
    \par
	\section{Background}
	\label{sec:format} 
	Before introducing our fragile watermarking method, we introduce the basic models in Sect 2.1, and a common malicious fine-tuning scenario is given in Sect 2.2.
	\subsection{Basic Model}
	\label{ssec:subhead}
	Most image recognition tasks are supervised learning and require sample data with corresponding labels in the stage of training classifier. Denote a batch of sample data with corresponding labels as: $\left\lbrace X,Y\right\rbrace  = \left\lbrace x_{i},y_{i}\right\rbrace_{i=1}^{N}$, where $N$ is size of batch, $X$ and $Y$ respectively represent sample data and corresponding labels. For a $k$ classification task, the label value $y\in\left\lbrace0,...,k-1 \right\rbrace.$ Suppose there is a classification model trained on sample data $D_{train}=\left\lbrace X_{train},Y_{train}\right\rbrace$, and this model is named as $C:X\to Y$, we refer to $C$ as the base classification model.
    \par
	Generative Adversarial Nets (GAN) \cite{goodfellow2014generative} consists of generative model $G$ and discriminative model. 
	Denote $\left\lbrace z_i\right\rbrace_{i=1}^n$ as random noise vectors, then $\left\lbrace z_i\right\rbrace_{i=1}^n$ can be mapped to $\left\lbrace x_i\right\rbrace_{i=1}^n$ through $G$. 
	In this paper, we use a generative model $G$ to generate fragile sample data $X_{trigger}$, where $X_{trigger}$ has the same shape as sample data of $D_{train}$.
	
	\subsection{Backdoor Attack}
	\label{ssec:subhead}
	Based on the basic classification model $C$, an adversary first preset backdoor \cite{gu2017badnets} data $X_{backdoor}$ which has the same data distribution as training data $D_{train}$, and there is no difference between these backdoor data and training data in human eyes. Hence, this adversary can fine-tune the model $C$ with $\left( X_{train}\cup X_{backdoor}\right) $ to inject backdoor data. Denote $C'$ as the backdoor model after fine-tuning which nearly has the same performance as $C$, but $C'$ recognizes backdoor data as preset results and model performance declines sharply if the adversary uses $X_{backdoor}$ as input data.
	\par
		\begin{figure}[t]
			\begin{minipage}[b]{1.0\linewidth}
		\centering
			\centerline{\includegraphics[width=8.5cm]{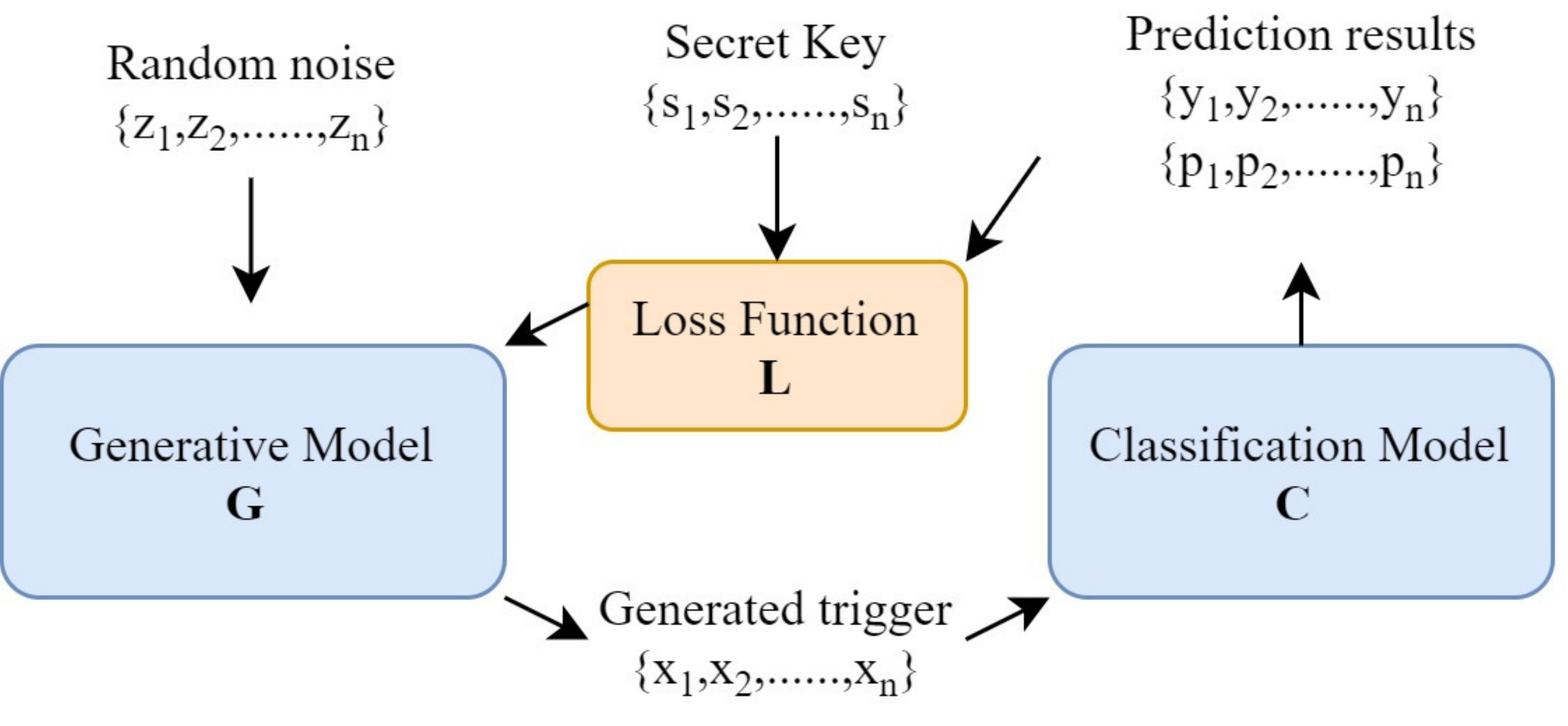}}
			\end{minipage}
		\caption{The framework of generating fragile trigger set.}
		\label{fig:res}
	\end{figure}
    \par	
	\section{Fragile watermarking}
	\label{ssec:pagestyle}
	\subsection{Overview}
	\label{ssec:subhead}
	Let's consider an application scenario, and there are three parties: model provider, model user, and adversary. The model provider uploads a trained model to cloud for serving, an adversary may use backdoor model to instead the original one. In the situation that without detail of pre-trained model from cloud, the model user wants to verify if the model served by model provider is actually the one he uploaded. Hence, we introduce a black-box fragile watermarking method without model performance degradation to verify the integrity of watermarked models. 
	\par
	Our idea is that the model provider first generates a fragile watermarking trigger set to mark $C$ before uploading trained $C$ to cloud.
	The model provider utilizes a generative model with a secret key $Y_{sk}=\left\lbrace s_{i}\right\rbrace_{i=1}^n$ to generate these fragile samples, whose labels predicted by $C$ are the same as $Y_{sk}$. 
	Then, the model provider can send generated fragile samples and $Y_{sk}$ to the model user directly. 
	The model user inputs these samples into the provided model from cloud and obtains the output $Y$. 
	The model integrity can be verified by comparing $Y$ and $Y_{sk}$.  
	\par
	\subsection{Watermarking Methodology}
	\label{ssec:subhead}	
	Fig. 1 shows our proposed fragile watermarking framework to mark the target classification model. 
	We first generate $n$ random noise vectors $\left\lbrace z_i\right\rbrace_{i=1}^n$ whose size is $1\times512$ as inputs of generative model and a secret key $Y_{sk}=\left\lbrace s_{i}\right\rbrace_{i=1}^n$. Then, $\left\lbrace z_i\right\rbrace_{i=1}^n$ are input into generative model $G$ to get generated fragile samples $X_{fragile}=\left\lbrace x_i\right\rbrace_{i=1}^n$. 
	Next, we input $X_{fragile}$ into the target classification model $C$ to get output results $Y_{fragile}=\left\lbrace y_{i}\right\rbrace_{i=1}^n$, and $\left\lbrace p_{i}\right\rbrace_{i=1}^n$. 
	$Y_{fragile}$ is the predicted category of generated fragile samples, and $\left\lbrace p_{i}\right\rbrace_{i=1}^n$ are predicted outcomes for each generated fragile sample after softmax operation. 
	At the first epoch, $Y_{fragile}$ is different from $Y_{sk}$, and $X_{fragile}$ is insensitive to target classification model modification too. 
	Hence, generative model $G$ is optimized with the loss function $L$ until $Y_{fragile}$ is the same as $Y_{sk}$, the composition of $L$ is listed as follows:
	\begin{itemize}
		\item $L_{cla}\left(Y_{fragile},Y_{sk}\right):$ $L_{cla}$ is the cross entropy loss, we adopt it to calculate the distance between $Y_{fragile}$ and $Y_{sk}$.
		\item $Var\left( P\right):$ This item is to calculate the variance of $P=\left\lbrace p_{i}\right\rbrace_{i=1}^n$, we add this regularization term in the stage of training $G$ to make $X_{fragile}$ more sensitive to model modification.
	\end{itemize} 
	Then, the $L$ loss funciton is writed as :
	\begin{equation}
		L = L_{cla}\left(Y_{fragile},Y_{sk}\right) + a\cdot Var\left( P\right)    
	\end{equation}
	$a$ is a weight coefficient, larger $a$ can improve the sensitivity of generated samples. 
	In the verification phase, we denote $AccTri$ as the authentication metrics to calculate the difference between predicted results of generated samples and $Y_{sk}$. The value of $AccTri$ can verify whether the classifier model has been modified, the $AccTri$ is defined as follows:
	\begin{equation}
		AccTri = \frac{\sum_{i=1}^n \mathbf{1}_{condition}\left( C\left( x_{i}\right)=y_i\right)}{n}
	\end{equation} 
	In $AccTri$, $\mathbf{1}_{condition}$ is a conditional function that returns 1 if the condition is true else returns 0. 
	If $AccTri < 1.0$, it indicates that the watermarked classifier has been modified.
    \par
	\section{Experiments}
	\label{sec:typestyle}
	Experiments consist of three subsections verifying our contributions.
	In Table 1, we compare recent fragile watermarking methods with our approach, and Table 2 further illustrates the difference between watermarking scheme \cite{zhu2021fragile} and ours in impacting the performance of the original model.
	The proposed watermarking method is utilized in the following subsections to obtain triggers and watermark the target classifier.
	Resnet\cite{he2016deep} is adopted as the default classifier $C$, and the last layer of classifiers refers to conv5\_x, average pool, and fully connected layer in \cite{he2016deep}. 
	PGAN\cite{karras2017progressive} is utilized as default generative model $G$. 
	In the watermarking stage, we preset a secret key $Y_{sk}$ as $\left\lbrace s_{i}\right\rbrace_{i=1}^n=\left\lbrace i\%M\mid i=0,1,...,n-1\right\rbrace$, $M$ is the total number of categories in train dataset, and the number of generated fragile triggers is $100$. 
	There are 60 epochs of fine-tuning the watermarked classifier in the verifying stage. 
	We adopt Stochastic Gradient Descent \cite{QIAN1999145} to fine-tune Resnet18 \cite{he2016deep} with a learning rate of $1e-3$ and Adam \cite{da2014method} to fine-tune larger resnet with a learning rate of $1e-5$.
	Since there is no modification to model $C$ in the proposed fragile watermarking method, there is no impact on the performance of model $C$.
	\par
    \begin{table}[htbp]
	    \centering
	    \caption{Comparison of recent fragile model watermarking methods with our approach.}
	    \vspace{3pt}
	    \setlength{\tabcolsep}{1.5mm}{
	    \begin{tabular}{ccccc}
	        \hline
	        Method & \makecell{Method\\Type} & \makecell{Model\\Accuracy} & \makecell{Trigger\\Generation} \\
	        \hline
	        \cite{guan2020reversible}-ACMMM 2020 & White-box & $\downarrow$ & N/A \\
	        \cite{botta2021neunac}-INS 2021 & White-box & $=$ & N/A \\
	        \cite{he2019sensitive}-CVPR 2019 & Black-box & $=$ & White-box  \\
	        \cite{xu2020secure}-ACSAC 2020 & Black-box & $=$ & White-box  \\
	        \cite{zhu2021fragile}-KSEM 2021 & Black-box & $\downarrow$ & Black-box  \\
	        Ours & Black-box & $=$ & Black-box  \\
	        \hline
	    \end{tabular}
	    }
	    \label{tab:my_label}
	\end{table}
	\vspace{-0.15cm}
    \begin{table}[htbp]
	    \centering
	    \caption{Impact comparison results of \cite{zhu2021fragile}-KSEM 2021 and our model watermarking method on the performance of the original model, in which the original model is Resnet18 \cite{he2016deep}, and the training dataset is CIFAR10 \cite{Krizhevsky_2009_17719}.}
	    \vspace{3pt}
	    \setlength{\tabcolsep}{1.2mm}{
	    \begin{tabular}{ccc}
	        \hline
	        Model & Accuracy & Difference  \\
	        \hline
	        Original Model & $91.2\%$ & N/A\\
	        Watermarked Model\cite{zhu2021fragile} & $90.3\%$ & $-0.9\%$\\
	        Watermarked Model(ours) & $91.2\%$ & $0$\\
	        \hline
	    \end{tabular}
	    }
	    \label{tab:my_label}
	\end{table} 
	\vspace{-0.15cm}
	\par
	\subsection{Ablation Study on Generating Trigger}
	\label{ssec:subhead}
	This part ablates the components of loss function $L$ to evaluate the effect on training generative model $G$. 
	We train three generators $G_{cla}, G_{full}, G_{Var}$ with different components of $L$. 
	$G_{cla}$ is trained with $L_{cla}$ and $Y_{sk}$, $G_{full}$ adopts full loss function $L$ and $Y_{sk}$ which weight $a$ is $200$, $G_{Var}$ is just trained with regularization term $Var$ and weight $a$ is $200$, $G_{non}$ is the initial generative model without training. 
	Except for $G_{non}$, all of them are trained for 300 epochs. 
	We adopt Resnet18 as default classifier $C$ which is trained on CIFAR10. 
	Table 3 shows that prediction probabilities of one or two categories are much higher than others when input generated triggers from $G_{non}$ and $G_{cla}$.  
	For generated triggers from $G_{full}$ and $G_{Var}$, almost half of the categories have very similar prediction probabilities.
	Thus, an additional regularization term $Var$ helps generate triggers in which more categories have similar prediction probability.
	\par
    \begin{table}[h]
	    \centering
	    \caption{The softmax prediction probabilities of categories for four generated triggers, and each trigger is randomly selected from the corresponding generative model. 
	    $'-'$ is adopted to instead corresponding values which are less than $1\ast10^{-2}$.}
	    \vspace{3pt}
	    \setlength{\tabcolsep}{0.7mm}{
	    \begin{tabular}{ccccccccccc}
	        \hline
	        & 0 & 1 & 2 & 3 & 4 & 5 & 6 & 7 & 8 & 9\\
	        \hline
	        $G_{non}$ & $.015$ & $-$ & $-$ & $.975$ & $-$ & $-$ & $-$ & $-$ & $-$ & $-$ \\
	        \hline
	        $G_{cla}$ & $-$ & $-$ & $-$ & $-$ & $-$ & $.998$ & $-$ & $-$ & $-$ & $-$\\
	        \hline
	        $G_{full}$ & $.149$ & $.146$ & $.146$ & $.148$ & $-$ & $.245$ & $-$ & $-$ & $.014$ & $.147$\\
	        \hline
	        $G_{var}$ & $-$ & $-$ & $.201$ & $.184$ & $.211$ & $-$ & $.216$ & $-$ & $-$ & $.182$\\
	        \hline
	    \end{tabular}
	    }
	    \label{tab:my_label}
	\end{table}  
	\subsection{Trigger Set Sensitivity Test}
	\label{ssec:subhead}
	Next, we examine the sensitivity of trigger set from above generative models.
	Generated triggers from $G_{non}$ and $G_{Var}$ are first input into $C$ to get corresponding initial labels. 
	Then, we adopt BIM \cite{kurakin2018adversarial} method to transform original training samples into backdoor data, utilize original training data with these backdoor data to fine-tune the classifier $C$. 
	In Fig.2, $AccTri$ of generated triggers from $G_{non}$, $G_{cla}$, $G_{full}$, and $G_{var}$ in each epoch is respectively recorded by Line a, b, c, and d. 
	As can be seen, Line c drops rapidly in the first epoch of fine-tuning and is much lower than Line a, b in the whole process. 
	It means that generated samples from $G_{full}$ are much more sensitive than $G_{non}$ and $G_{cla}$. 
	Hence, our proposed loss function in training generative models is valid to generate a fragile trigger set sensitive to model fine-tuning. 
	\par
	
	\begin{figure}[t]
	\begin{minipage}[b]{1.0\linewidth}
		\centering
		\centerline{\includegraphics[width=8.5cm]{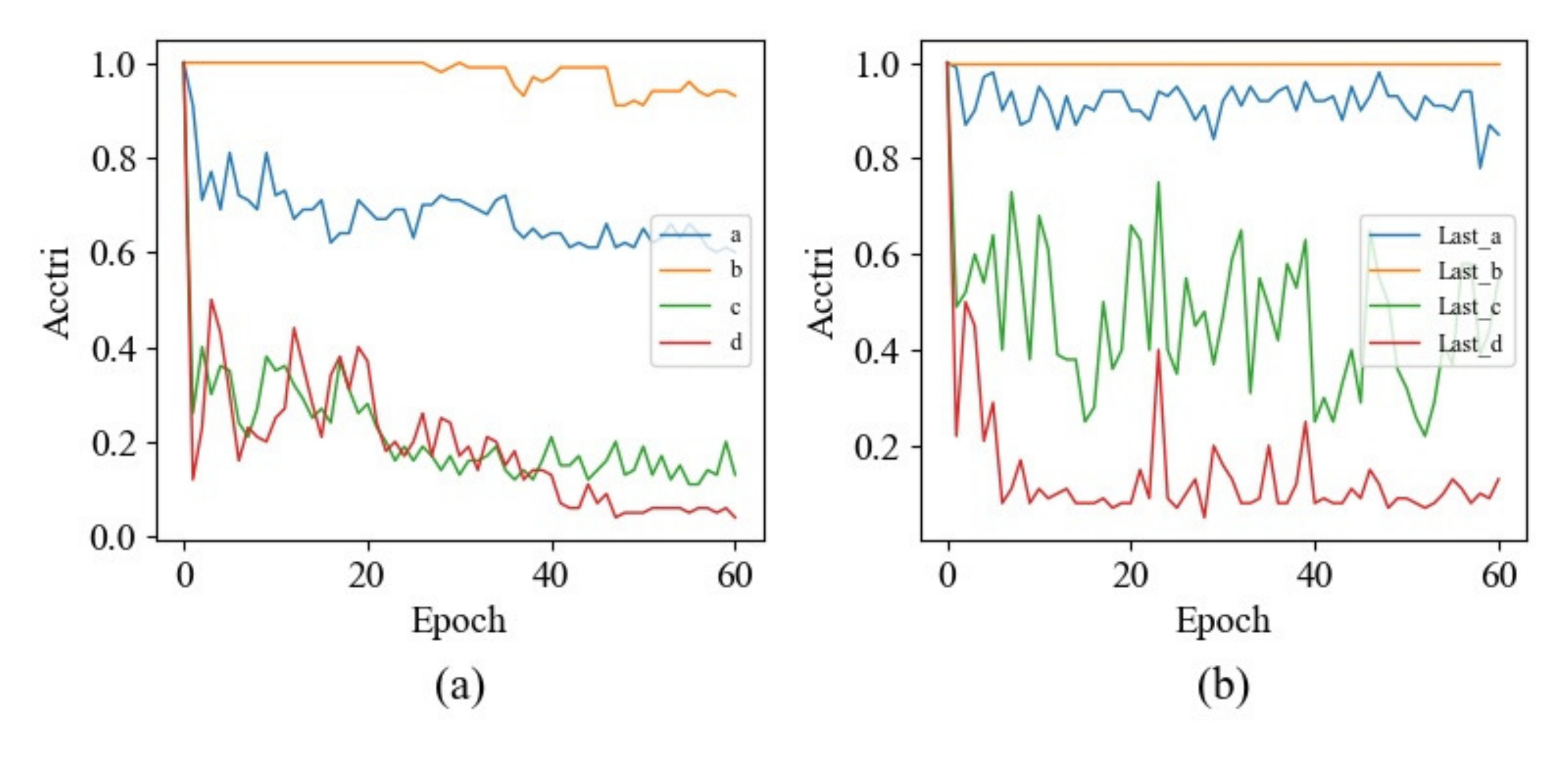}}
	\end{minipage}
    \vspace{-1.0cm}
	\caption{The $AccTri$ of generated samples from $G_{non}$, $G_{cla}$, $G_{full}$, and $G_{var}$ in each test epoch. 
	Experimental verifications are carried out in (a) where all model parameters can be modified and (b) where only the last model layer is modified.}
	\label{fig:res}
    \end{figure}
	\par
    \subsection{Trigger Set Extensibility Test}
	\label{ssec:subhead}
	In this part, we evaluate whether increasing parameters of classification models impact the sensitivity of generated fragile watermarking triggers. 
	Therefore, Resnet18, Resnet50 \cite{he2016deep}, Resnet101 \cite{he2016deep}, and Resnet152 \cite{he2016deep} are selected as default classification model for testing, only the last layer of model parameters can be modified. 
	Different weight coefficient $a$ is used to train generative models with full loss function. 
	Table 4 records the mean $AccTri$ values of fragile watermarking triggers generated from these generative models in the whole test process. 
	If weight $a$ is set as $1$, the $AccTri$ is not decline in the first epoch of fine-tuning.
	In this situation, our fragile watermarking method may have omissions for detecting fine-tuning of the target classifier.
	When weight $a$ exceeds or equals $100$, the $AccTri$ declines sharply in the first epoch of fine-tuning. 
	And larger weight $a$ can decrease the mean $AccTri$, the lower $AccTri$ means that fine-tuned classification models misclassify more trigger samples. 
	Hence, the generator trained with larger $a$ can get a fragile trigger set which is more sensitive to model modification, and our model fragile watermarking method is effective for larger classification models.
	\par
    \begin{table}[t]
	    \centering
	    \caption{The mean $AccTri$ of fragile samples generated with larger weight $a$ and classification models. The corresponding item is crossed if $AccTri$ has not changed in each fine-tuning epoch.}
	    \vspace{3pt}
	    \setlength{\tabcolsep}{1.0mm}{
	    \begin{tabular}{ccccc}
	        \hline
	        & Resnet-18 & Resnet-50 & Resnet-101 & Resnet-152 \\
	        \hline
	        $a=1$ & $\times$ & $\times$ & $\times$ & $\times$ \\
	        $a=100$ & $0.56$ & $0.46$ & $0.37$ & $0.47$ \\
	        $a=200$ & $0.43$ & $0.38$ & $0.31$ & $0.38$ \\
	        $a=400$ & $0.34$ & $0.32$ & $0.26$ & $0.33$ \\
	        $a=800$ & $0.26$ & $0.25$ & $0.23$ & $0.27$ \\
	        \hline
	    \end{tabular}
	    }
	    \label{tab:my_label}
	\end{table}  
	We also consider the possible impact from increased categories of training dataset and then train Resnet18 with CIFAR100 \cite{Krizhevsky_2009_17719} as default classification model $C$.
	The total number of categories increased from 10 to 100. 
	A generative model $G$ is trained with weight value $4000$, and only the last layer of default classification model is fine-tuned in testing. 
	In 60 epochs of fine-tuning model $C$, the max $AccTri$ of generated fragile trigger set is $0.97$ and the mean is $0.68$, the $AccTri$ still declines in the first epoch of fine-tuning.
	It shows that the generated fragile trigger set is still sensitive to modificaiton of classification models with 100 predicted categories. 
	\par


	\section{Conclusion and future work}
	\label{sec:ref}
	In this paper, we propose a novel neural network fragile watermarking with no model performance degradation. 
	In our approach, watermarking process of classifier and training process of generative model are done simultaneously, there is no detail and modification of the target classifier are required in the whole process of watermarking. 
	We demonstrate that generators trained with proposed loss effectively get fragile trigger sets sensitive to model fine-tuning and our approach is compatible to watermark enormous classifiers.
	In the future, we will incorporate an explainable framework as part of the model watermarking evaluation and comparison. 
	
	\bibliographystyle{IEEEbib}
	\bibliography{strings,refs}
	
\end{document}